\documentclass[letterpaper, 10 pt, conference]{ieeeconf} 
\makeatletter
\renewcommand{\@IEEEsectpunct}{ \,}
\IEEEoverridecommandlockouts 
\usepackage{graphicx}
\usepackage{amsmath}
\usepackage{amssymb}
\usepackage{mathtools}
\usepackage{dsfont}
\usepackage{hyperref}

\title{\LARGE \bf
Self-Supervised Instance Segmentation by Grasping
}

\author{YuXuan Liu$^{1,2}$,
Xi Chen$^1$, 
Pieter Abbeel$^{1,2}$%
\thanks{$^{1}$Covariant.ai, $^{2}$University of California, Berkeley.}%
}

\begin{document}

\maketitle
\thispagestyle{empty}
\pagestyle{empty}

\begin{abstract}

Instance segmentation is a fundamental skill for many robotic applications.
We propose a self-supervised method that uses grasp interactions to collect segmentation supervision for an instance segmentation model. 
When a robot grasps an item, the mask of that grasped item can be inferred from the images of the scene before and after the grasp.
Leveraging this insight, we learn a grasp segmentation model to segment the grasped object from before and after grasp images. 
Such a model can segment grasped objects from thousands of grasp interactions without costly human annotation.
Using the segmented grasped objects, we can ``cut" objects from their original scenes and ``paste" them into new scenes to generate instance supervision.
We show that our grasp segmentation model provides a 5x error reduction when segmenting grasped objects compared with traditional image subtraction approaches.
Combined with our ``cut-and-paste" generation method, instance segmentation models trained with our method achieve better performance than a model trained with 10x the amount of labeled data.
On a real robotic grasping system, our instance segmentation model reduces the rate of grasp errors by over 3x compared to an image subtraction baseline.
\end{abstract}

\IEEEpeerreviewmaketitle

\begin{figure}
    \centering
    \includegraphics[width=\linewidth]{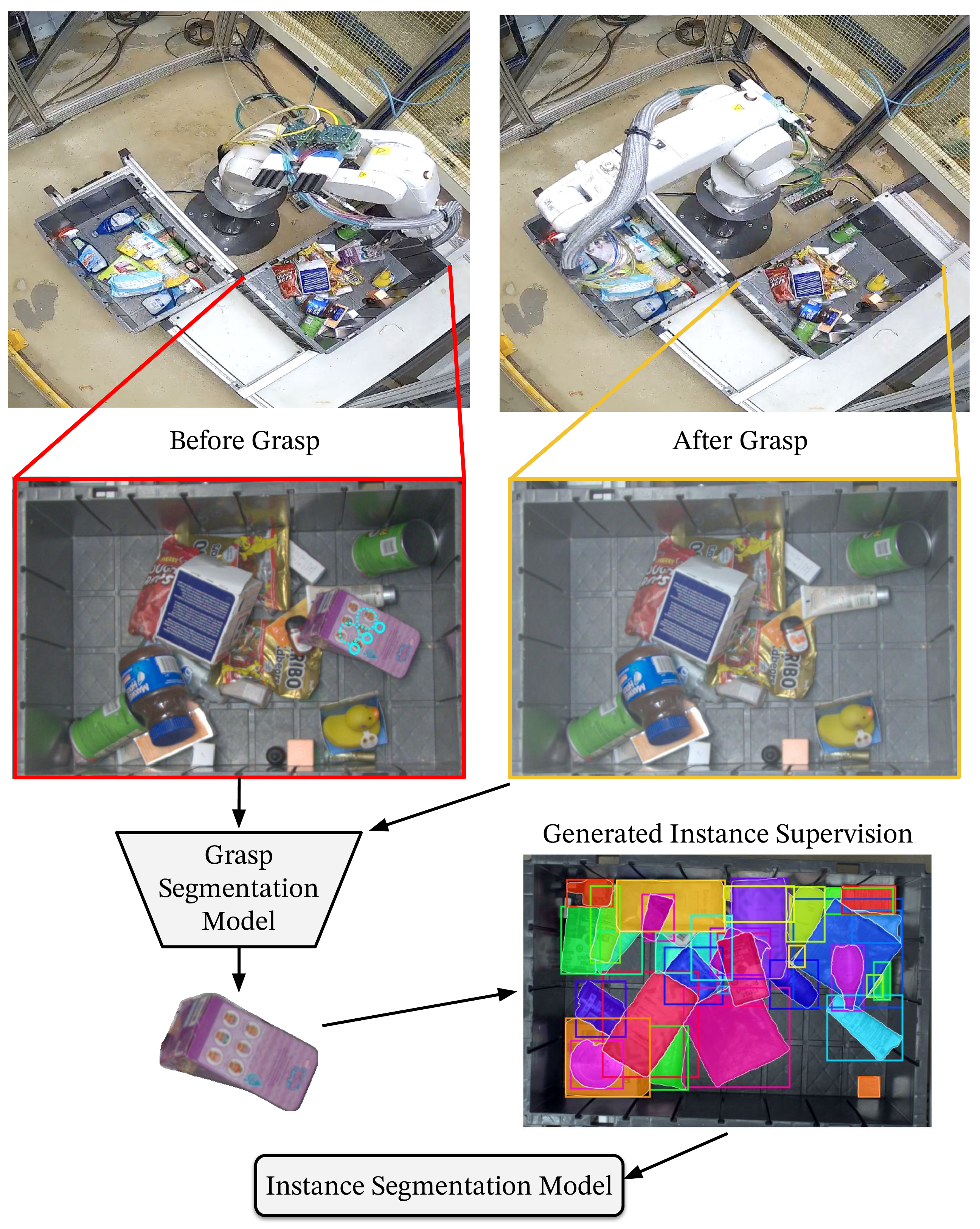}
    \caption{Overview of our method: the grasp segmentation model takes before and after images of the grasp to predict the mask of the grasped object. 
    We use the grasp segmentation model to get object masks for thousands of grasps in a self-supervised manner. 
    These grasped objects can be ``cut" and ``pasted" to generate diverse training supervision for the instance segmentation model. }
    \label{fig:overview}
\end{figure}

\section{Introduction}
Instance segmentation is often the basis of many robotic applications, including object grasping, manipulation, and placement.
Given an image, the goal of instance segmentation is to predict the set of pixels that belong to each object.
After objects are detected, a robot can use segmentation masks to execute more generalizable grasping and manipulation policies. A robust robotic application must be able to recognize thousands of new objects on an ongoing basis. 

Much of the recent advances in instance segmentation \cite{mask2former, maskrcnn, maskdino, latent-mrcnn} assume that large-scale labeled datasets of objects from known classes are available \cite{coco, pascalvoc}.
This assumption, however, does not hold for many robotics applications that must handle a constant stream of new objects.
Collecting and annotating such a dataset is also costly and expensive.
How can a robot learn to segment a diverse range of objects with only limited labeled data?

An object can be defined as a contiguous group of pixels that move together \cite{eisen}.
When the robot successfully grasps an object, the pixels of that object are removed from the scene.
The mask of the grasped object can then be inferred from the grasp location, before image, and after image of the scene.
Leveraging this insight, we propose a grasp segmentation model that predicts the mask of the grasped object.
A grasp segmentation model differs from an instance segmentation model in that it only needs to predict one object (the grasped one), and has additional information (before/after images and the grasp location) that enables it to generalize better even when trained on a small dataset. 

Traditional methods of segmenting grasp interactions often use image and background subtraction which are not robust to occlusions, reflections, and other objects moving \cite{seg_by_int}. 
We find that our learned model overcomes these limitations and can robustly segment grasped objects, even when other objects in the scene have shifted during grasping.
In our experiments, we show that our grasp segmentation model is significantly more accurate at segmenting grasped objects than traditional image subtraction approaches.

Once we have a grasp segmentation model, we can run this on thousands of unlabeled grasps to get self-supervised object masks.
To reduce noise, we use a suction gauge to keep only successful grasps and propose an uncertainty-aware filtering method to keep high-accuracy masks.
With these filtered object masks, we can then ``cut" objects from their original scenes and ``paste" them into new scenes with random augmentations to generate instance supervision \cite{copy_paste_gan, paste_long_tail}.
We combine this generation scheme with inpainting to generate a diverse set of photo-realistic scenes with infinite variation of scale, rotation, and occlusion.
This allows our self-supervised robotic system to continually learn to segment new objects, and improve on known objects, without human annotation.
Figure~\ref{fig:overview} shows an overview of our self-supervised instance segmentation method.

The key contributions of this paper are as follows:
\begin{itemize}
\itemsep0em 
    \item [1.] We propose a self-supervised robotic grasping system that can continually learn and improve its instance segmentation, on new and known objects, without human annotation.
    \item [2.] Our novel grasp segmentation model uses before and after grasp images to segment grasped objects with 5x less error than traditional approaches, while being robust to occlusions, reflections, and other object moving in the scene.
    \item [3.] We introduce a ``cut-and-paste" and inpaint method to generate supervision for instance segmentation models that outperforms the same model trained with 10x the amount of labeled data.
    \item [4.] On a robotic grasping task, we show that models trained with our method can reduce the rate of grasping failures by over 3x compared to an image subtraction baseline.
\end{itemize}

\section{Related work}

\subsection{Instance Segmentation}
There has been a significant amount of recent advances in the field of instance segmentation, with many approaches focusing on supervised learning on large datasets such as COCO \cite{coco}.
Detect-then-segment is among the earliest learned approaches that use a two-stage object detection and segmentation architecture \cite{maskrcnn}.
More recently, single-stage models that use transformer attention mechanisms, such as Mask2Former, have demonstrated strong performance \cite{mask2former, maskformer, detr, swin}.
However, all of these methods rely on the availability of large-scale annotated datasets, which may be costly to obtain for robotics applications that must handle a constant stream of new objects.

\subsection{Self-Supervised Segmentation}
To address this issue, there have been several approaches that propose self-supervised or semi-supervised methods for instance segmentation.
Cut and paste methods have been shown to improve instance segmentation performance \cite{copy_paste_gan, paste_long_tail}. 
Prior work \cite{seg_by_int, learn_interact} proposed self-supervised approaches that use image subtraction on before and after grasps to provide instance segmentation supervision. 
However, masks recovered from naive image subtraction are imperfect, resulting in worse performance at higher IOU thresholds, which is insufficient for high-performing robotic applications.
Optical flow methods can also infer contiguous groups of pixels that move together as the robot is pushing them around \cite{self_instance_interact, eisen, singulate_and_grasp}. 
This may be an impractical approach if continuous high-bandwidth video is not available from the robot camera, or if the robot is expected to be grasping objects instead of pushing them in a production environment.

\subsection{Moving Object Detection}
Another class of methods can detect moving objects in video sequences, such as traffic and surveillance footage. 
These methods can use a background subtraction approach by modeling a static background scene with a mixture of Gaussians \cite{mog, mog2}. 
Pixels that are noticeably different from the background model, as determined at the pixel level or with local features \cite{lsbp}, are segmented as moving objects.
More recently, convolutional neural networks have been applied to learn this background subtraction with 2D and 3D convolutions \cite{deep_background_sub, 3d_background_sub}. 
Moving object detection models, however, are not directly applicable for grasp segmentation since they segment all objects that have moved instead of the one that was grasped. 
If the robot moves an adjacent object to the grasped object, both objects would be segmented as one under moving object detection, providing incorrect instance segmentation supervision.
Moreover, methods that learn a background model will have limited data since the background scene changes with every grasp.

\subsection{Representation Learning}
Other approaches have proposed learning object \textit{representations} for robotics instead of instance segmentation directly.
In Grasp2Vec \cite{grasp2vec}, representations of the object and scene are learned to satisfy arithmetic consistency. 
These representations can then be used to learn policies that manipulate and grasp objects.
Similarly, pixel-wise descriptors can be learned for each object using a contrastive loss \cite{dense_descriptors}. 
While learned object representations can be used for robot manipulation, they have yet to be proven effective for learning instance segmentation. 
The clear semantics of instance segmentation may be desirable for some robotic applications such as counting the number of objects, or ensuring that grasps only occur on a single object.

\section{Instance Segmentation by Grasping}

\begin{figure*}
    \centering
    \includegraphics[width=0.85\linewidth]{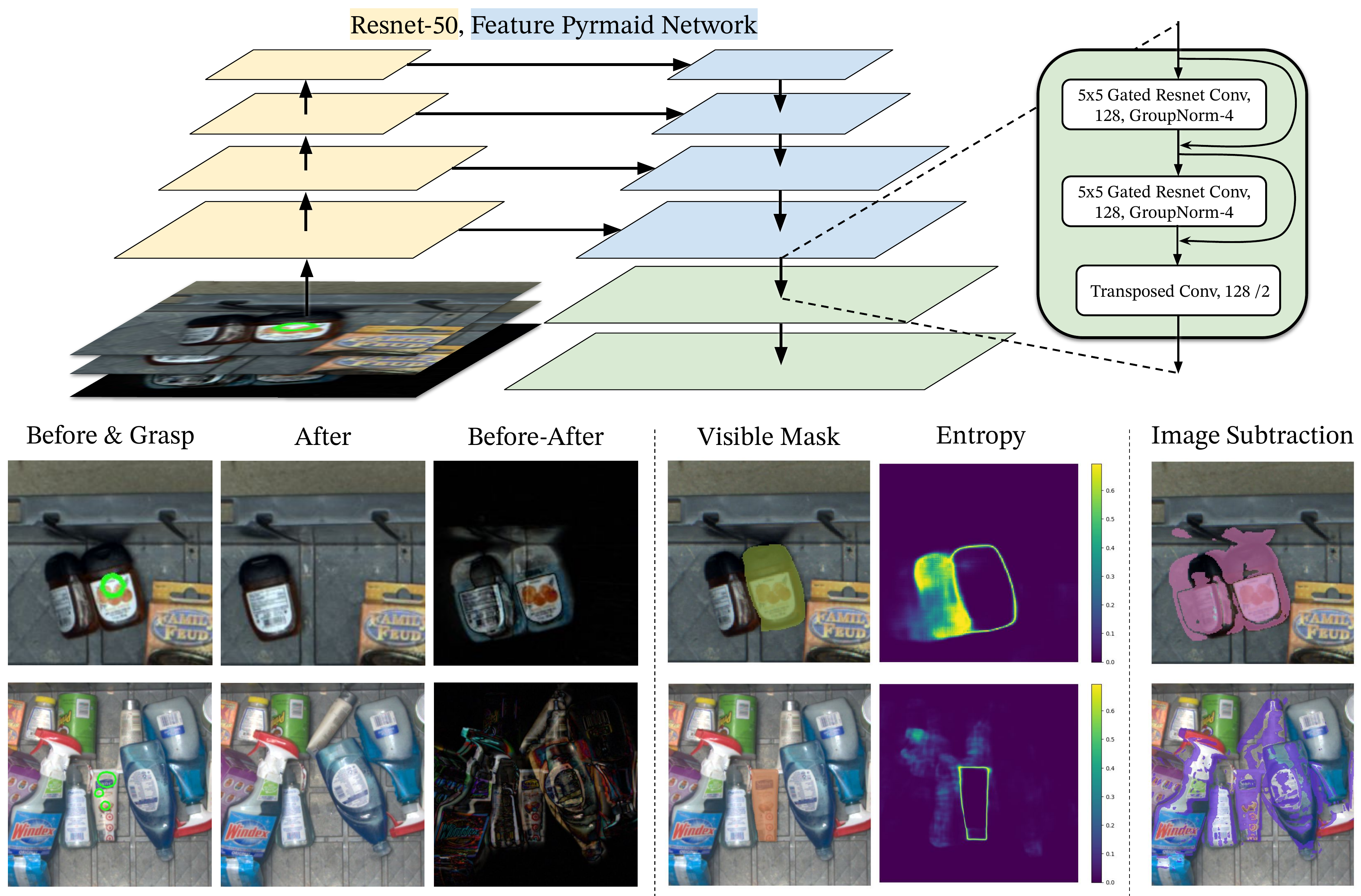}
    \caption{Our grasp segmentation model takes before, after, grasp, and subtraction images as input to predict the mask of the grasped object. We build upon a Resnet-50 backbone with a Feature Pyramid Network (FPN) and two additional learned upsampling convolution blocks to predict masks at the image resolution. Traditional image subtraction methods (on the right) fail to properly handle reflections and other objects moving in the scene. Our learned approach can correctly segment the grasped object and express uncertainty via the entropy of the prediction.}
    \label{fig:grasp-segm}
\end{figure*}

Our method uses grasp interactions to collect segmentation data for objects in a self-supervised manner.
We propose a grasp segmentation model that can robustly segment grasped objects from before and after grasp images.
We show how our model can predict masks that are robust to occlusions, reflections, and other objects in the scene changing.
Then, by combining our grasp segmentation model with our uncertainty-aware filtering method, we can collect a dataset of grasped object masks from unlabeled grasp images.

\subsection{Grasp Segmentation Model}\label{sec:grasp-segm}
Our grasp segmentation model is inspired by prior work that uses grasp interactions to infer object masks and representations based on sets of pixels that move together \cite{seg_by_int, grasp2vec}. 
Given a before grasp image $i_b$, an after grasp image $i_a$, and a grasp mask $g$, our model predicts the visible mask of the grasped object $m_v$.
This is similar to image subtraction which subtracts the before and after images $|i_b-i_a| > t$ and uses a threshold $t$ to determine the object mask. 
However, image subtraction is a very brittle segmentation method with very few tunable parameters that limit its robustness to occlusions, reflections, and other moving objects.

Our model, on the other hand, uses a neural network to predict both the visible $m_v$ and amodal mask $m_a$ (including occlusions) \cite{amodal} of the grasped object.
By explicitly reasoning for occlusions, we can filter out objects that would have been partially visible and providing incorrect supervision for downstream instance segmentation.
A trained model will also learn to ignore inputs, such as reflections and other moving objects, that are not relevant to segmenting the grasped object.

We base the grasp segmentation model architecture on a Resnet-50 \cite{resnet} with a Feature Pyramid Network (FPN) \cite{fpn}, which has been shown to be effective at segmenting objects at multiple scales. 
To initialize the model with features that are amenable for object detection, we load weights from a Mask-RCNN \cite{maskrcnn} pre-trained on COCO \cite{coco}.
We include image subtraction $i_b-i_a$ as an input to the model since it provides a good inductive bias for which pixels have changed.
Altogether, we concatenate the inputs $[i_b, i_a, i_b-i_a, g]$ along the channel dimension before passing them into the Resnet-50 backbone.
Since the input dimensions are different than the usual RGB inputs for Resnet, we randomly initialize the first layer of the Resnet-50 while initializing all other weights from the pre-trained Mask-RCNN.

To make predictions at the original input resolution, we take the highest resolution, stride-4 feature layer from the FPN, and apply a series of convolutions and transposed convolutions to upsample the features.
We use two blocks of gated residual convolutions followed by a transposed convolution for upsampling. 
Similar to regular residual blocks $y = x + C_1(x)$, gated residual blocks use an additional learned gating operation $y = x + \sigma (C_2(x))C_1(x)$ where $\sigma$ is the sigmoid function and $C$ are learned convolution operators, like in LSTMs \cite{lstm}.
For our application, this enables the model to easily ignore irrelevant features such as using the reflective object features to mask out the predicted object mask.
Finally, we make a 2-channel prediction $\hat{m}_v$, $\hat{m}_a$ corresponding to the visible and amodal object masks respectively. 
Figure~\ref{fig:grasp-segm} illustrates our model architecture overview along with sample inputs and predictions on how our model can be more robust than image subtraction.

\begin{figure*}
    \centering
    \includegraphics[width=0.9\linewidth]{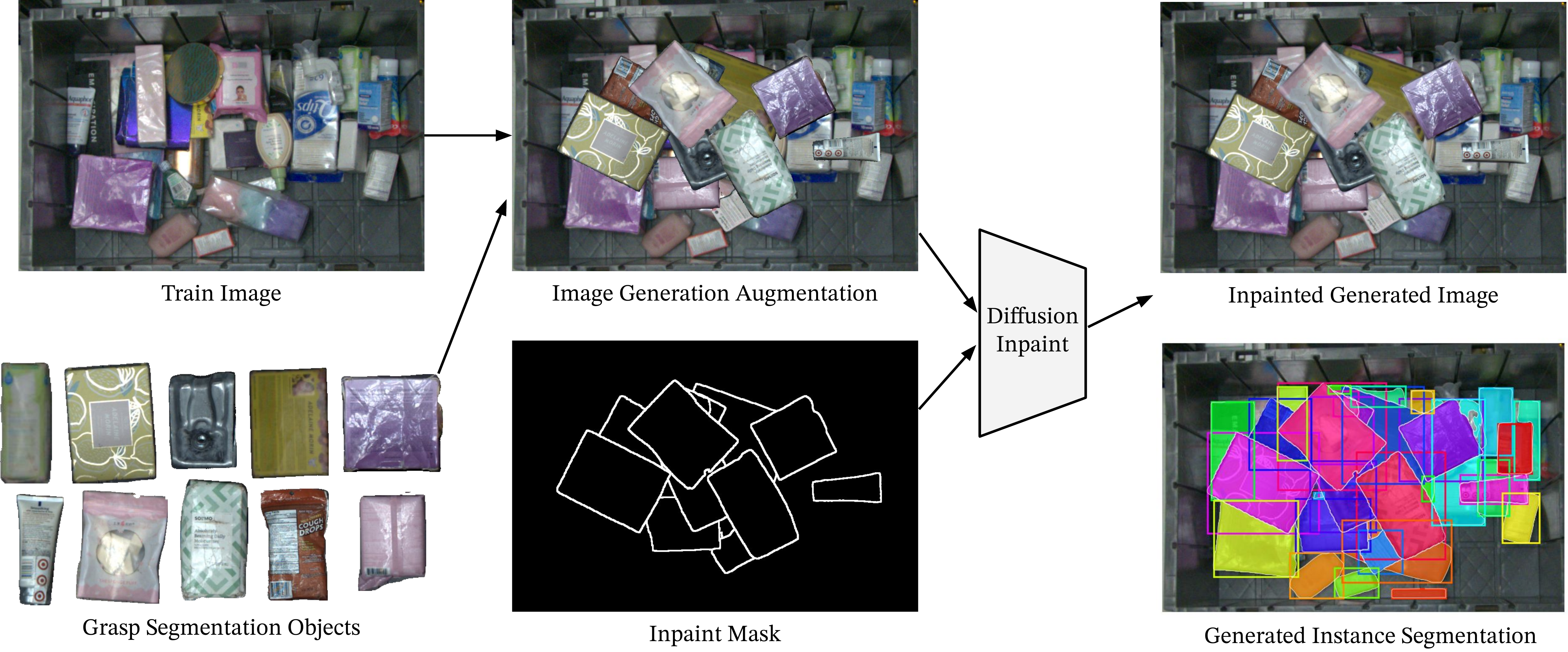}
    \caption{By applying the grasp segmentation model to thousands of grasps, we can collect a large dataset of grasp segmentation objects in a self-supervised manner. 
    Then we can generate instance segmentation supervision by taking a random subset of segmented grasped objects and ``pasting" them onto any training image. 
    We apply augmentations such as rotation, scale, and offset to generate cluttered scenes of objects.
    Then, we use a pre-trained diffusion inpaint model to smooth out the pasted object boundaries to be more photorealistic.
    The resulting inpainted image and pasted masks can be used to train any off-the-shelf instance segmentation model.}
    \label{fig:inpaint-generation}
\end{figure*}

We supervise all predictions using a binary cross entropy loss with weighting $w$ 
\begin{equation}
CE(\hat{m}, m, w) = \frac{1}{n} \sum_{i=1}^n w_i(m_i \log \hat{m}_i + (1-m_i) \log (1-\hat{m}_i)) 
\end{equation}
Due to the imbalanced nature of the classification task, we use multiple weightings to ensure the neural network gives the appropriate attention to the relevant pixels.
The first two weights, $w^{(1)} = m, w^{(2)} = \neg m$ provide a balanced weighting. 
Since it's important for the pixels near the mask boundary to be accurate, we use $w^{(3)...(6)} = maxpool(m, k)$ with kernel sizes $k=(11, 51, 101, 201)$. 
This enlarges the region around the object and focuses the model's loss to predict an accurate object boundary.

We train with the total loss
\begin{equation}
    L(\hat{m}, m) = \sum_{j=1}^6 CE(\hat{m}, m, w^{(j)})
\end{equation}
using Adam with learning rate 5e-6 with a batch size of 8 until convergence.
We train the grasp segmentation model using supervised learning on a small dataset of 100-200 labeled grasp images. 
To improve generalization, we use augmentations during training including random crops, resizes, blurs, and color adjustments.
Even with such a small training set, the model can generalize to provide accurate segmentation masks for thousands of grasps on new objects.

\subsection{Supervising Instance Segmentation}

Once our grasp segmentation model is trained, how can we learn an instance segmentation model?
We propose a 4-step approach: first, we use the grasp segmentation model to collect and filter object masks across a large variety of grasped objects. 
Then, we ``cut-and-paste" object masks using augmentations to generate cluttered scenes of objects.
We use inpainting on object boundaries to reduce pasting artifacts, and use the inpainted images to train a state-of-the-art instance segmentation model.
Figure~\ref{fig:inpaint-generation} illustrates an overview of our approach.

\subsubsection{Collecting Accurate Object Masks:}\label{sec:filter-object-masks}
First, we use the grasp segmentation model on unlabeled grasp image pairs to predict the mask of the grasped object.
To collect a reliable set of unoccluded object crops, we apply a several filters on the prediction.
We compute the sums of visible and amodal masks $S_v = \sum \hat{m}_v, S_a = \sum \hat{m}_a$, and compare the ratio $\frac{S_v}{S_a} > t_{occ}$ with a threshold. We use a threshold $t_{occ}=0.95$ to filter out objects the model predicts as not fully visible.

The grasp segmentation model could also predict masks that are wrong or uncertain. To filter out uncertain masks, we can use entropy thresholding where 
\begin{equation}
    \mathcal{H}(\hat{m}) = -\hat{m}\log \hat{m} - (1-\hat{m}) \log(1-\hat{m})
\end{equation}
is the binary entropy. We can threshold average relative entropy of the visible mask prediction $\frac{\sum\mathcal{H}(\hat{m}_v)}{\sum\hat{m}_v} < t_{ent}$
to filter out uncertain predictions. We found $t_{ent} = 0.1$ to provide a reasonable trade-off between precision and recall.

Finally, predicting discontinuous masks can indicate that more than one object was detected or the model is predicting spurious blobs. 
To avoid this kind of prediction error, we use OpenCV's findContours function to find contiguous groups of contours. 
Then we count the number of contours with at least 1000 pixels and only keep objects with 1-2 contours.

\subsubsection{Image Generation Augmentation:}\label{section:paste}
Once we have filtered out uncertain predictions and obtained a set of accurate object masks, we can use these masks to generate supervision for instance segmentation. 
We use object masks to ``cut" images of the objects and ``paste" them onto annotated images to generate instance supervision \cite{copy_paste_gan, paste_long_tail}.

To train the model to be robust to a diverse set of objects, we randomize select 0-25 objects to be pasted on a random training image.
We apply different geometric transformations to the objects such as randomly rotating up to 360 degrees, scaling between 0.75x-1.25x, and randomly selecting a position. 
This can help the model learn to better handle variations in the object orientations, positions, and occlusions in the images.
With a small amount of labeled data, we can augment with self-supervised grasp segmentation to generate a large dataset of instance segmentation supervision

\subsubsection{In-Painting With Diffusion:}

While naive cut-and-paste can generate diverse supervision, the augmented images may contain artifacts such as wrong object boundaries and unrealistic shadows.
These artifacts can lead a model to learn features that are not present in natural images. 
To bring the generated images closer to the natural image space, we use a pre-trained Stable Diffusion in-painting model that is trained on a large-scale image dataset \cite{stable_diffusion}.

First, we take the visible boundary of the objects that are pasted and dilate the boundary by 5 pixels on all sides to cover the region on the object boundary.
We use this mask as the in-paint mask input to a pre-trained Stable Diffusion in-painting model and run the denoising process for 4 steps on the masked region. 
This provides a more photo-realistic boundary between the background and the object. 

\subsubsection{Instance Segmentation Model:}\label{section:instance-model}

Since our method directly generates instance segmentation supervision, it's compatible with any off-the-shelf instance segmentation model.
This enables greater flexibility of our method since we can take advantage of any past and future advances in model architecture.
We chose a state-of-the-art model, Mask2Former \cite{mask2former}, as the main instance segmentation model for all of our experiments.
Mask2Former uses a multi-scale, masked-attention transformer decoder with learned queries and a Hungarian matching loss on the masks.
We use the official public implementation of Mask2Former with a Resnet-50 as the backbone and default parameters (100 queries). 

\section{Evaluating Grasp Segmentation}

First, we'll evaluate the performance of our grasp segmentation model compared to traditional methods such as image subtraction.
Since grasp segmentation provides object masks for instance segmentation supervision, the accuracy of the grasp segmentation will influence the performance of downstream instance segmentation.

\subsection{Grasp Data Collection}\label{section:dataset}

For our grasping experiments, we use an ABB-1200 robot with a 5 suction cup end-effector as shown in Figure~\ref{fig:overview}.
The robot uses RGB-D cameras to plan pick and place motions that cycle the objects between two bins.
We record the image (640x960 pixels) before the grasp $i_b$, the image after the grasp $i_a$, and the grasp mask $g$ corresponding to the pixels of the active suction cups.
After grasping, we use suction gauges of each cup to determine whether the object was indeed picked and whether the grasp was successful.
While this data collection method requires an instance segmentation model with decent performance to generate grasps, we use the suction gauge to keep only successful grasps in the dataset which can bootstrap even a poorly performing model.

We collect 110k grasp image pairs across a variety of training and test objects. 
We label 1k images from the training set and 6k images from the test objects with instance segmentation labels.
This enables us to train and evaluate both grasp and instance segmentation. 
The test objects and the training objects are distinct and have no overlap.

\subsection{Background and Image Subtraction}

In order to evaluate the performance of our grasp segmentation model, we will compare it to traditional methods such as image and background subtraction \cite{seg_by_int, mog, mog2, lsbp}. Image subtraction is a simple method that subtracts the before-grasp image from the after-grasp image to obtain the grasp mask. This method assumes that changed pixels correspond to the grasped object.
To make the image subtraction more robust, we apply greyscale and Gaussian blur before the subtraction, $G(I) = GaussianBlur(Greyscale(I))$. 
Then we use the thresholded difference as the segmentation mask $$\hat{m} = |G(I_a) - G(I_b)| > t$$

We also consider background subtraction methods such as mixture of gaussians MOG \cite{mog}, MOG2 \cite{mog2} and Local SVD binary pattern (LSBP) \cite{lsbp}.
MOG models the background with a mixture of gaussians while LSBP uses local image features to detect changes. 
With all image and background subtraction methods, we can apply the same OpenCV contour approximation from Section~\ref{sec:filter-object-masks} as a filter. 

\begin{table}[t]
  \centering
  \caption{Grasp Segmentation Evaluation}\label{table:grasp-segmentation}
  \begin{tabular}{|l|c|c|c|}
    \hline
    Method & mIOU & Error & Recall \\
    \hline
    Subtract & 51.1 & 396\% & 98.6\\
    Subtract-Filter & 61.6 & 44.5\% & 50.4 \\
    MOG & 51.8 & 333\% & 98.3\\
    MOG-Filter & 59.9 & 47.1\% & 65.4\\
    MOG2 & 32.4 & 1289\% & 99.8\\
    MOG2-Filter & 47.3 & 970\% & 42.0\\
    LSBP & 37.7 & 1050\% & 99.8\\
    LSBP-Filter & 55.6 & 359\% & 44.0\\
    \hline
    Grasp-100-NoAug & 78.3 & 31.5\% & 100 \\
    Grasp-100 & 78.7 & 22.6\% & 99.8 \\
    Grasp-100-NoAug-Filter & 80.4 & 26.9\% & 92.3 \\
    Grasp-200 & 86.8 & 14.3\% & 99.6 \\
    Grasp-100-Filter & 91.1 & 9.05\% & 29.4 \\
    Grasp-200-Filter & \textbf{92.6} & \textbf{7.63\%} & 63.9 \\
    \hline
  \end{tabular}
\end{table}

\subsection{Evaluation Results} \label{section:grasp-segmentation-eval}

\bgroup
\def\arraystretch{1.1}
\begin{table*}[t]
  \centering
  \caption{Instance Segmentation Evaluation}\label{table:instance-segmentation}
  \begin{tabular}{|l|l|l|ccc|cc|}
    \hline
    Method & Paste From & Labeled Images & AP & AP@0.5 & AP@0.75 & AP$^L$ & AP$^M$\\
    \hline
    Mask2Former-100 & None & 100 & 49.30 & 70.75 & 51.47 & 55.34 & 41.70\\
    Single-Object-100 & None & 100 & 22.10 & 24.05 & 23.22 & 27.03 & 15.48\\
    Paste-Subtract & Subtract-Filter & 100 & 62.11 & 80.13 & 67.01 & 69.97 & 52.83\\
    Paste-Subtract-Robust & Subtract-Filter  & 100 & 65.07 & 86.24 & 72.51 & 72.97 & 53.10\\
    Paste-Train & Train Objects & 100 & 66.28 & 84.22 & 70.54 & 73.27 & 55.88\\
    Paste-Grasp & Grasp-100  & 100 & 69.68 & 87.37 & 75.37 & 77.45 & 58.47\\
    Paste-Grasp-Filter & Grasp-100-Filter  & 100 & 69.22 & 86.15 & 74.62 & 77.63 & 56.72\\
    Paste-Grasp-Robust & Grasp-100-Filter & 100 & 63.02 & 85.33 & 70.63 & 70.47 & 51.54 \\
    Paste-Grasp-Inpaint & Grasp-100 + Inpaint  & 100 & \textbf{72.33} & \textbf{88.25} & \textbf{78.11} & \textbf{80.68} & 59.40\\
    \hline
    Mask2Former-1000 & None  & 1000 & 70.08 & 85.69 & 74.34 & 76.12 & 61.48\\
    \hline
  \end{tabular}
\end{table*}
\egroup

We train the Grasp Segmentation model on a subset of 100 and 200 grasps from the training set, until convergence as described in Section~\ref{sec:grasp-segm}.
Then we evaluate on the test set of hold-out object grasps, using mean intersection over union (mIOU) as the evaluation metric. 
We also report the relative error rate, which is the number of incorrectly predicted pixels divided by the number of pixels in the ground truth mask. 
For methods that use filtering described in Section~\ref{sec:filter-object-masks}, we calculate the recall, which is the percentage of mask predictions not filtered out. 
When using filtering, the mIOU and Error rates are calculated only on the kept data after filtering. 
If a method makes no pixel predictions for a grasp, this prediction will be filtered as well.

\subsubsection{How Do Our Methods Compare With Baselines?}
The results of our evaluation are summarized in Table~\ref{table:grasp-segmentation}. 
All grasp segmentation models perform better than image and background subtraction baselines (Subtract, MOG, MOG2, LSBP). 
As shown in Figure~\ref{fig:grasp-segm}, these traditional approaches are unable to account for other objects moving in the scene and will confuse pixels that change with similar intensity.

\subsubsection{How Does Performance Vary With Training Set Size?}
We can see that the grasp segmentation model with filtering trained on 200 grasps, Grasp-200-Filter, performs the best. 
Training with 100 grasps and filtering (Grasp-100-Filter) achieves slightly worse mIOU, error rate, and less than half the recall of Grasp-200-Filter).
This suggests that increasing the amount of training data improves the accuracy and the number of scenes the model can confidently segment.

\subsubsection{What's The Impact Of Filtering And Augmentations?}
Filtering improves the mIOU and error rates while decreasing recall for all methods. 
We also see that not using data augmentation, Grasp-100-NoAug-Filter, achieves lower accuracy metrics but higher recall. 
This suggests that fewer objects are filtered out and the model is confidently wrong since it has overfit to the small training set.
Data augmentation is important to training a robust grasp segmentation model.

\section{Evaluating Instance Segmentation}

To evaluate instance segmentation performance, we use the Mask2Former model with R50 backbone and default hyperparameters as described in Section~\ref{section:instance-model}.
We train and evaluate on the dataset from Section~\ref{section:dataset} and use early stopping to prevent overfitting. 
We use 100 labeled training images for most evaluations, while benchmarking on 1000 labeled training images as a reference.
We evaluate instance segmentation models using standard metrics, such as overall Average Precision (AP), IOU thresholded AP (AP@0.5, AP@0.5), and object size breakdowns (AP$^L$, AP$^M$).

For Paste methods, we take the corresponding model trained in Section~\ref{section:grasp-segmentation-eval} to generate object crops.
We apply the same inference and filtering for each method on the 110k unlabeled grasp dataset. 
Using the filtered object crops, we then paste them randomly onto the training set following the procedure described in Section~\ref{section:paste}.
As an ablation, we also paste from training image crops to evaluate our augmentation scheme without grasp data (Paste-Train).

\subsection{Baseline Methods}
\subsubsection{Single Object Supervision: } 
One intuitive way to use grasp segmentation data for instance segmentation is to supervise only on the grasped object.
With the Mask2Former loss, this can be done by using only the predicted grasp segmentation in the Hungarian matching and ignoring the "no object" loss term.
We combine single-object supervision from the Grasp-100 model with full supervision from the training set at a 50:50 ratio (Single-Object-100). 

\subsubsection{Robust Set Loss:}
To overcome errors in noisy masks generated by image subtraction, prior work \cite{seg_by_int} proposed a robust set loss, which requires the predicted mask to be within only a margin of the ground truth mask.
Given a predicted mask, the robust set loss will use a discrete optimization to find the closest target mask that is within some IOU margin with the ground truth.
We use the publicly available implementation of robust set loss with IOU threshold 0.7 and replace the Mask2Former mask loss with the robust set loss.

\subsection{Evaluation Results}\label{section:instance-results}

\begin{figure*}
    \centering
    \includegraphics[width=0.9\linewidth]{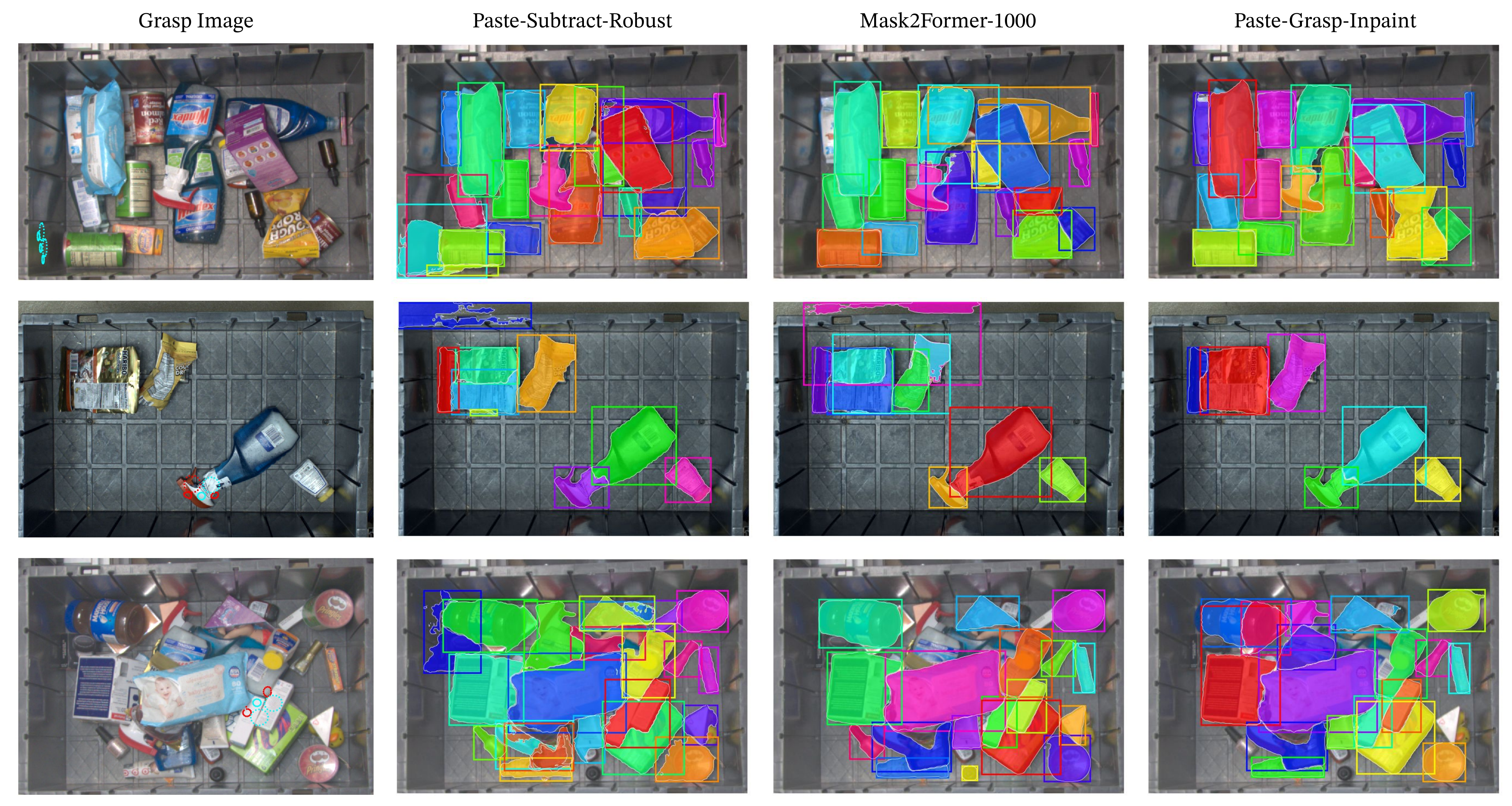}
    \caption{Visualizations of different types of grasp failures from real robot execution caused by wrong segmentation. 
    The first column shows the executed grasp with active cups colored in blue and inactive cups colored in red.
    Columns 2-4 show the predicted segmentation from each of the 3 models used in the evaluation.
    The first row shows an example of a grasp on a non-object that was incorrectly predicted by the segmentation model.
    The second row is a grasp on an unstable part of the object that leads to a drop; the segmentation model splits the object into two masks, causing an unstable grasp to be executed.
    The third row is a grasp on two objects; this is caused by the segmentation model grouping two objects into one mask.
    }
    \label{fig:segm-comparison}
\end{figure*}

\subsubsection{How Do Baseline Methods Compare?}
Table~\ref{table:instance-segmentation} summarizes the results of our evaluation.
Training with single object supervision, Single-Object-100, actually results in worse performance than just training on the full supervision dataset (Mask2Former-100). 
We suspect that there isn't enough negative supervision from the single-object data since there are only labels for a single object in each image.
We find that pasting with objects from filtered image subtraction (Paste-Subtract), provides a performance improvement over just supervised learning (Mask2Former-100). 
Adding the robust set loss (Paste-Subtract-Robust) further improves the segmentation performance which is consistent with \cite{seg_by_int}.
Pasting from the labeled training objects, Paste-Train, outperforms Paste-Subtract-Robust on AP.
This suggests that the accuracy of cropped objects in the training set outweighs the diversity of the less correct crops from image subtraction. 
Moreover, pasting provides a strong augmentation for learning rotational, scale, and occlusion invariant instance segmentation.

\subsubsection{Does Robust Set Loss Always Improve Performance?}
Using the robust set loss with object masks from the more accurate, grasp segmentation model (Paste-Grasp-Robust) actually hurts performance, unlike in the image subtraction case.
This suggests that the robust set loss is beneficial only when there is a significant amount of error in the grasp segmentation, such as with image subtraction's 44.5\% error.
With the 9.05\% error rates of our grasp segmentation model, supervising with a normal cross entropy loss is better.

\subsubsection{What's The Impact Of Filtering?}
Using object masks from our learned Grasp Segmentation model (Paste-Grasp) can further improve instance segmentation performance. 
Paste-Grasp and Paste-Grasp-Filter have similar results which suggest there is some trade-off between the error of the object mask and how many objects are filtered out.
Recall from Table~\ref{table:grasp-segmentation} that Grasp-100 achieves 22.6\% error with 99.8 recall, while Grasp-100-Filter has 9.05\% error with 29.4 recall. 
The non-filtered model (Paste-Grasp) performs slightly better which suggests that the diversity of the objects seen outweighs the accuracy of the object segmentation here.

\subsubsection{How Does Inpainting Affect Performance?}
Finally, we see that using inpainting with the grasp segmentation model (Paste-Grasp-Inpaint), achieves the best performance, outperforming a Mask2Former model trained on 10x the amount of labeled images on all but one metric. 
This suggests that inpainting the pasted objects can produce more realistic supervision for learning instance segmentation features.

\section{Robot Grasping Evaluation}

\subsection{Experimental Setup}
To evaluate the effectiveness of our method in a real robotic application, we use our trained instance segmentation models to detect objects for grasping.
Using the same grasping setup from Section~\ref{section:dataset} and the models trained in Section~\ref{section:instance-results}, we compare grasping success with different instance segmentation models.
A grasp attempt is counted as a success if it picks up exactly one object and places it into the other bin without dropping.
If the predicted mask is not an actual object, or contains multiple objects, this would be a grasp failure.
The grasping system will try to land as many cups as possible on a detected object mask and plan a collision-free path to reach that object.
By keeping all other parts of the grasping system constant and only changing the segmentation model, we can isolate the effects of the segmentation model on grasp performance.

\begin{table}[t]
  \centering
  \caption{Robot Evaluation Results}\label{table:robot-metrics}
  \begin{tabular}{|l|c|c|}
    \hline
    Method & Grasp Error Rate \\
    \hline
    Paste-Subtract-Robust & 29.02\% \\
    Mask2Former-1000 & 8.78\% \\
    Paste-Grasp-Inpaint & \textbf{8.22\%} \\
    \hline
  \end{tabular}
\end{table}

\subsection{Evaluation Results}

We compare the top segmentation models from each method: supervised learning (Mask2Former-1000), robust image subtraction (Paste-Subtract-Robust), and our grasp segmentation model with inpainting (Paste-Grasp-Inpaint).
We perform 700 grasps on the same object set for each segmentation model.
The results of our robotic grasping evaluation are shown in Table~\ref{table:robot-metrics}. 

Overall we found the grasping evaluation to be consistent with the instance segmentation evaluation in Section~\ref{section:instance-results}.
Paste-Subtract-Robust performs the worst with most of its grasp failures due to predicting masks that don't belong to any object, such as reflections in the wall or empty areas of the bin.
Our Paste-Grasp-Inpaint model achieves the lowest grasp error rate that is over 3x better than the image subtraction baseline and comparable to a model trained with 10x the amount of labeled data, Mask2Former-1000. 

\subsection{Failure Analysis}

Figure~\ref{fig:segm-comparison} highlights three failure categories caused by incorrect instance segmentation that we will discuss here.

\subsubsection{Grasp On Non-Objects:} 
In the first row, the robot grasps on the reflection in the bottom left corner of the bin, which fails since it's not on an object. 
Grasping on non-objects such as reflections and empty areas of the bin is a common failure of the Paste-Subtract-Robust model.
We suspect image subtraction may have incorrectly segmented the reflection or bin as an object, which then provided misguided supervision to the instance segmentation model.
Our learned model, on the other hand, is robust to these errors and does not make the same mistakes.

\subsubsection{Splitting Objects Into Two:} 
In the second row, an unstable grasp on the top part of the bottle leads to the robot dropping the item. 
Here the segmentation model incorrectly split one object into two masks, resulting in a grasp that is not centered on the object.
All three models fail to segment the bottle correctly, however our learned Paste-Grasp-Inpaint model has the least amount of errors on other objects.

\subsubsection{Grouping Two Object As One:} 
Finally, when the segmentation model incorrectly groups two objects as one, this can lead to a grasp on two objects.
Only the Paste-Subtract-Robust model suffers from this failure case, which suggests that image subtraction incorrectly grouped two objects that moved as one object.
This incorrectly grouped object was then used to supervise the model, leading to this grasp error.

\section{Conclusion} 
\label{sec:conclusion}

In this work, we proposed a novel method for instance segmentation that utilizes self-supervised grasp images to generate object masks for training. 
We showed that our grasp segmentation model can accurately detect objects with high mIOU and low error rate, even when trained on a small number of labeled images. 
We then use the grasp object masks to train an instance segmentation model with a inpainting augmentation method, which outperforms a model trained with 10x the amount of labeled data. 
We have also shown that our method leads to improved grasping performance in a real-world robotic application. 
This work highlights the potential of using self-supervised grasp images for learning instance segmentation models, and opens up new possibilities for training such models in a wide range of robotic applications.

\bibliographystyle{IEEEtran}
\bibliography{references}

\end{document}